\documentclass[twoside,11pt]{article}

\usepackage[preprint]{jmlr2e}

\let\proof\relax

\newcommand{\dev}{\textit{d}}

\usepackage{amsmath,amsfonts,bm}

\def\1{\bm{1}}

\DeclareMathAlphabet{\mathsfit}{\encodingdefault}{\sfdefault}{m}{sl}
\SetMathAlphabet{\mathsfit}{bold}{\encodingdefault}{\sfdefault}{bx}{n}

\newcommand{\cmark}{\ding{51}}%
\newcommand{\xmark}{\ding{55}}%

\usepackage{microtype}
\usepackage{booktabs} 
\usepackage{amsfonts,amsmath,amsthm,amssymb,color,graphicx}
\usepackage{enumerate}
\usepackage{physics}
\usepackage{multirow}
\usepackage{multicol}
\usepackage{enumitem}
\usepackage{mathtools}
\usepackage{caption}
\usepackage{pifont}
\usepackage{subcaption}
\usepackage{tablefootnote}
\usepackage[dvipsnames]{xcolor}
\usepackage{soul}

\usepackage{lastpage}
\ShortHeadings{Survival Ordinary Differential Equation Networks}{Tang, Ma, Mei, and Zhu}

\firstpageno{1}

\begin{document}

\title{SODEN: A Scalable Continuous-Time Survival Model through Ordinary Differential Equation Networks}

\author{\name Weijing Tang\thanks{Equal contribution.} \email weijtang@umich.edu \\
       \addr Department of Statistics\\
       University of Michigan\\
       Ann Arbor, MI 48109, USA
       \AND
       \name Jiaqi Ma\footnotemark[1] \email jiaqima@umich.edu \\
       \addr School of Information\\
       University of Michigan\\
       Ann Arbor, MI 48109, USA
       \AND
       \name Qiaozhu Mei \email qmei@umich.edu \\
       \addr School of Information and Department of EECS\\
       University of Michigan\\
       Ann Arbor, MI 48109, USA
       \AND
       \name Ji Zhu \email jizhu@umich.edu \\
       \addr Department of Statistics\\
       University of Michigan\\
       Ann Arbor, MI 48109, USA
       }

\editor{Jie Peng}

\maketitle

\begin{abstract}
In this paper, we propose a flexible model for survival analysis using neural networks along with scalable optimization algorithms. 
One key technical challenge for directly applying maximum likelihood estimation (MLE) to censored data is that evaluating the objective function and its gradients with respect to model parameters requires the calculation of integrals. 
To address this challenge, we recognize from a novel perspective that the MLE for censored data can be viewed as a differential-equation constrained optimization problem. Following this connection, we model the distribution of event time through an ordinary differential equation and utilize efficient ODE solvers and adjoint sensitivity analysis to numerically evaluate the likelihood and the gradients. Using this approach, we are able to 1)~provide a broad family of continuous-time survival distributions without strong structural assumptions, 2)~obtain powerful feature representations using neural networks, and 3)~allow efficient estimation of the model in large-scale applications using stochastic gradient descent. Through both simulation studies and real-world data examples, we demonstrate the effectiveness of the proposed method in comparison to existing state-of-the-art deep learning survival analysis models. The implementation of the proposed SODEN approach has been made publicly available at \href{https://github.com/jiaqima/SODEN}{https://github.com/jiaqima/SODEN}.
\end{abstract}

\begin{keywords}
  Survival Analysis, Ordinary Differential Equation, Neural Networks
\end{keywords}

\section{Introduction}

Survival analysis is an important branch of statistical learning where the outcome of interest is the time until occurrence of an event, such as survival time until death and lifetime of a device until failure. In real-world data collections, some events may not be observed due to a limited observation time window or missing follow-up, which is known as \textit{censoring}. In this case, instead of observing an \textit{event time}, we record a \textit{censored time}, for example, the end of the observation window, to indicate that no event has occurred prior to it. 
Survival analysis methods take into account the partial information contained in the censored data and have crucial applications in various real-world problems, such as rehospitalization, cancer survival in healthcare, reliability of devices, and customer lifetime~\citep{chen2009study,miller2011survival,modarres2016reliability}.

Modern data collections have been growing in both scale and diversity of formats. For example, electronic health records of millions of patients over several decades are readily available, and they include laboratory test results, radiology images, and doctors' clinical notes. Work towards more flexible and scalable modeling of event times has attracted great attention in recent years. In particular, various deep neural network models have been introduced into survival analysis due to their ability in automatically extracting useful features from large-scale raw data ~\citep{faraggi1995neural, ching2018cox,katzman2018deepsurv,lee2018deephit,gensheimer2018simple,chapfuwa2018adversarial,kvamme2019time, https://doi.org/10.1002/sim.8542,  10.1093/bioinformatics/btaa1082}.

As a natural choice for estimating a probabilistic model, likelihood-based methods have been widely used for both traditional and deep survival analysis. However, a major challenge for scalable maximum likelihood estimation of neural network models lies in difficult-to-evaluate integrals due to the existence of censoring. Specifically, for an uncensored observation~$i$ whose event time $T=t_i$ is recorded, the likelihood is the \textit{probability density function} (PDF) $p(t_i)$. But, for a censored observation $j$, only the censored time $C=t_j$ is recorded while the event time $T$ is unknown. The likelihood of observation $j$ is the \textit{survival function} $S(t_j)$, which is the probability of no event occurring prior to $t_j$: $S(t_j) = \mathcal{P}\{T > t_j\} = 1 - \int_{0}^{t_j}p(s)ds.$
This integral imposes an intrinsic difficulty for optimization: evaluating the likelihood and the gradient with respect to parameters requires the calculation of integrals, which usually has no closed forms for most flexible distribution families specified by neural networks. 

To address this challenge, most existing works try to avoid the integrals in the following two ways: 1)~making additional structural assumptions so that no integral is included in the objective function, such as partial-likelihood-based methods under the proportional hazard (PH) assumption~\citep{cox1975partial}, or making parametric assumption that leads to closed-form integration in the likelihood~\citep{wei1992accelerated}; 2)~discretizing the continuous event time with pre-specified intervals so that the integral is simplified into a cumulative product. However, the structural and parametric assumptions are often restrictive and thus limit the flexibility of the model \citep{ng1997empirical, zeng2007maximum}; further, stochastic gradient descent algorithms cannot be directly applied to the partial-likelihood-based objective functions and thus limit the scalability of the model. As for discretization of the event time, it will likely cause information loss and introduce pre-specified time intervals as hyper-parameters.

In this paper, we recognize that maximizing the likelihood function for censored data can be viewed as an optimization problem with differential equation (DE) constraints, and thereby tackle the aforementioned optimization challenges with an efficient numerical approach. We propose to specify the distribution of event time through an ordinary differential equation (ODE) and utilize well-implemented ODE solvers to numerically evaluate the likelihood and its gradients. In particular, we consider the \textit{hazard function} $\lambda_x(\cdot)$\footnote[1]{The hazard function describes the instantaneous rate at which the event occurs given survival, and is a popular modeling target in survival analysis. Probabilistic meanings of the hazard, the cumulative hazard, and the likelihood form in terms of the hazard and cumulative hazard (see Eq. (\ref{general likelihood})) are shown in Section~\ref{prob}.} and its integral, the \textit{cumulative hazard function} $\Lambda_x(\cdot)$, in an ODE with a fixed initial value: 
\begin{equation}
\label{eq:ode}
\left\{
\begin{array}{lr}
\Lambda'_x(t) = h(\Lambda_x(t), t; x, \theta) \\
\Lambda_x(0) =0
\end{array}
\right.
,
\end{equation}
where the function $h(\cdot, \cdot, \cdot, \theta)$ is modeled by a neural network taking the cumulative hazard $\Lambda_x(t)$, the time $t$, and the feature $x$ as inputs and parameterized by $\theta$.  
Since the likelihood given both uncensored and censored data can be re-written in a simple form of the hazard and the cumulative hazard\footnotemark[1], we can evaluate the likelihood function by solving the above ODE numerically. Moreover, the gradient of the likelihood with respect to $\theta$ can be efficiently calculated via adjoint sensitivity analysis, which is a general method for differentiating optimization objectives with DE constraints~\citep{pontryagin1962mathematical, 10.1111/j.1365-246X.2006.02978.x}.  
We name the proposed method as SODEN, Survival model through Ordinary Differential Equation Networks.

In comparison to existing methods described above, the proposed SODEN is more flexible to handle event times allowing for a broad range of distributions without strong structural assumptions. Further, we directly learn a continuous-time survival model using an ODE network, which avoids potential information loss from discretizing event times. We empirically evaluate the effectiveness of SODEN through both simulation studies and experiments on real-world datasets, and demonstrate that SODEN outperforms state-of-the-art models in most scenarios. 

The rest of the paper is organized as follows. In Section~\ref{sec:background}, we provide a brief background on survival analysis and related work. In Section~\ref{sec:approach}, we describe the proposed model and the corresponding learning approach. We evaluate the proposed method using simulation studies in Section~\ref{sec:simulation} and on real-world data examples in Section~\ref{sec:real}. Finally Section~\ref{sec:conclusion} concludes the paper. 

\section{Background}
\label{sec:background}

In this section, we provide necessary preliminaries on survival analysis and summarize existing related work.

\subsection{Preliminaries}
\label{prob}

\subsubsection{The probabilistic framework of survival analysis} 
\label{sec: prob func}
Denote the non-negative event time by $T$ and the feature vector by $X$. We are interested in the conditional distribution of $T$ given $X=x$. In addition to the PDF, the distribution of $T$ can be uniquely determined by any one of the followings: the survival, the hazard, or the cumulative hazard function. We introduce definitions of these functions below. Denote the PDF by $p_x(\cdot)$ with $\int p_x(t)\dev t=1$. The survival function $S_x(\cdot)$ is the probability that no event occurred before time $t$, that is $S_x(t) =  \mathcal{P}\{T>t|X=x\}$. The hazard rate $\lambda_x(t)$ characterizes the instantaneous rate at which the event occurs for individuals that are surviving at time $t$, which is denoted by 
\[\lambda_x(t) = \lim_{\epsilon\rightarrow 0 }\frac{ \mathcal{P}\{t < T\leq t+\epsilon|T>t, X=x\}}{\epsilon} = \frac{p_x(t)}{S_x(t)}.\]
The cumulative hazard function $\Lambda_x(\cdot)$ is the integral of the hazard, that is $\Lambda_x(t) = \int_{0}^t\lambda_x(u) \dev u$. It follows that $S_x(t) = \exp(-\Lambda_x(t)) = \exp(- \int_{0}^t\lambda_x(u) \dev u)$. Thus, either the hazard function or the cumulative hazard function can specify the distribution of $T$. In particular, the hazard function $\lambda_x(\cdot)$ is a popular modeling target due to its practical meaning and informativeness in survival analysis. 

\subsubsection{Likelihood function} 
Below, we provide the likelihood for a family of distributions given independent identically distributed (i.i.d.) observations. We consider the common right-censoring scenario where the event time $T$ can be observed only if it does not exceed the censoring time~$C$. Let $Y=\min\{T,C\}$ indicate the observed time and $\Delta = \mathrm{1}_{\{T \leq C\}}$ indicate whether we observe the actual event time. We observe i.i.d. triplets $D_i=(y_i,\Delta_i, x_i)$ for $i=1,\cdots,N$. Under the standard conditional independence assumption of the event time and the censoring time given features, the likelihood function is proportional to
\begin{equation}
\label{general likelihood}
\prod_{i=1}^N p_{x_i}(y_i)^{\Delta_i}S_{x_i}(y_i)^{1-\Delta_i}=\prod_{i=1}^N \lambda_{x_i}(y_i)^{\Delta_i}e^{-\Lambda_{x_i}(y_i)},
\end{equation} 
where uncensored observations contribute the PDF and censored observations contribute the survival function. By definition, the likelihood function can also be written in terms of the hazard and the cumulative hazard as in (\ref{general likelihood}).

\subsection{Related Work}
\subsubsection{Traditional survival analysis} There has been a large body of classical statistical models dealing with censored data in the literature. The Cox model~\citep{cox1972regression}, which is probably the most commonly used model in survival analysis, makes the proportional hazard (PH) assumption where the ratio of the hazard function is constant over time. Specifically, the hazard function consists of two terms: an unspecified baseline hazard function and a relative risk function, that is 
\begin{equation}
\label{cox formula}
\lambda_x(t) = \lambda_0(t)\exp(g(x;\theta)).
\end{equation}
The Cox model also assumes that the relative risk linearly depends on features, that is $g(x;\theta) = x^T\theta$. In practice, however, either or both of the above assumptions are often violated. As a consequence, many alternative models have been proposed~\citep{aalen1980model, Buckley1979, Gray_1994, doi:10.1002/sim.4780020223, 10.1093/biomet/82.4.835, lin1995semiparametric, 10.1093/biomet/85.4.980, chen2002semiparametric, Shen2000, Wu2019Flex}.  Among them, to address the limitation of multiplicative hazard, a broader family that involves multiplicative and additive hazard rate has been proposed~\citep{aalen1980model, lin1995semiparametric}. To address the limitation of time-invariant effects, \citet{Gray_1994} has adapted the Cox model with time-varying coefficients to capture temporal feature effects. Alternatively, the accelerated failure time (AFT) model assumes that the logarithm of the event time is linearly correlated with features~\citep{Buckley1979,wei1992accelerated}, that is $\log T= x^T\theta+\epsilon$. When the error $\epsilon$ follows a specific parametric distribution such as log-normal and log-logistic, the likelihood in (\ref{general likelihood}) under AFT model has a closed-form and can be efficiently optimized. Although the aforementioned models are useful, they often model the effect of features on the survival distribution in a simple, if not linear, way. These restrictions prevent the traditional models from being flexible enough to model modern data with increasing complexity.

\subsubsection{Deep survival analysis} 
There has been an increasing research interest on utilizing neural networks to improve feature representation in survival analysis. Earlier works~\citep{faraggi1995neural, ching2018cox, katzman2018deepsurv} adapted the Cox model to allow nonlinear dependence on features but still make the PH assumption. For example, \citet{katzman2018deepsurv} used neural networks to model the relative risk $g(x;\theta)$ in (\ref{cox formula}). \citet{kvamme2019time} further allowed the relative risk to vary with time, which resulted in a flexible model without the PH assumption. Specifically, they extended the relative risk as $g(t,x;\theta)$ to model interactions between features and time. These models are all trained by maximizing the partial likelihood~\citep{cox1975partial} or its modified version, which does not need to compute the integrals included in the likelihood function.  The partial likelihood function is given by 
\begin{equation}
\label{pl}
\mathcal{PL}(\theta;D) = \prod_{i:\Delta_i=1}\frac{\exp(g(y_i, x_i;\theta))}{\sum_{j\in R_i}\exp(g(y_i, x_j;\theta))},
\end{equation}
where $R_i = \{j: y_j \geq y_i\}$ denotes the set of individuals who survived longer than the $i^{th}$ individual, which is known as the \textit{at-risk} set. Note that evaluation of the partial likelihood for an uncensored observation requires access to all other observations in the at-risk set. 
Hence, stochastic gradient descent (SGD) algorithms cannot be directly applied to partial likelihood-based objective functions, which is a serious limitation in training deep neural networks for large-scale applications. 
In the worst case, the risk set can be as large as the full data set. When the PH assumption holds, i.e., the numerators and denominators in~(\ref{pl}) do not depend on $y_i$, evaluating the partial likelihood has a time complexity of $O(N)$ by computing $g(x_i;\theta)$ once and storing the cumulative sums. For flexible non-PH models, under which the likelihood has the form as (\ref{pl}), the time complexity further increases to $O(N^2)$. 
Although in practice one can naively restrict the at-risk set within each mini-batch, there is a lack of theoretical justification for this ad-hoc approach and the corresponding objective function is unclear.

On the other hand, SGD-based algorithms can be naturally applied to the original likelihood function. Following this direction, \citet{lee2018deephit} and \citet{gensheimer2018simple} propose to discretize the continuous event time with pre-specified intervals, such that the integral in (\ref{general likelihood}) is replaced by a cumulative product.
This method scales well with large sample size and does not make strong structural assumptions. However, determining the break points for time intervals is non-trivial, since too many intervals may lead to unstable model estimation while too few intervals may cause information loss. 

\begin{table}[!t]
  \centering
  \begin{tabular}{lcccc}
    \toprule
    Model & Non-linear & No PH Assumption & Continuous-time & SGD \\
    \midrule
    Cox  & \textcolor{gray}{\xmark}  & \textcolor{gray}{\xmark} & \cmark & ?\footnotemark[1] \\
    DeepSurv & \cmark & \textcolor{gray}{\xmark} & \cmark & ?\\
    DeepHit & \cmark & \cmark & \textcolor{gray}{\xmark} & \cmark\\
    Nnet-survival & \cmark & \cmark & \textcolor{gray}{\xmark} & \cmark\\
    Cox-Time & \cmark &  \cmark  & \cmark & ?\\
    SODEN (proposed) & \cmark & \cmark & \cmark & \cmark \\
    \bottomrule
  \end{tabular}
  \caption{Comparison between the proposed method, SODEN, and related work, Cox~\citep{cox1972regression}, DeepSurv~\citep{katzman2018deepsurv}, DeepHit~\citep{lee2018deephit}, Nnet-survival~\citep{gensheimer2018simple}, and Cox-Time~\citep{kvamme2019time}.}
  \label{Table: relatedwork}
\end{table}

We note that there are works that also consider a continuous event time but they do not optimize the likelihood function. Instead, they target summary statistics of the event time distribution such as the restricted mean survival time or the survival probability at a fixed time point \citep{https://doi.org/10.1002/sim.8542,  10.1093/bioinformatics/btaa1082}. During the review process, we became aware of an independent and concurrent related work~\citep{groha2020neural}, which proposes a neural-network-based ODE approach to model the Kolmogorov forward  equation  that  characterizes  the  transition  probabilities  for  multi-state  survival analysis.

The proposed SODEN is a flexible continuous-time model and is trained by maximizing the likelihood function, where SGD-based algorithms can be applied.
Table~\ref{Table: relatedwork} summarizes the comparison between SODEN and several representative existing methods.

\footnotetext[1]{SGD algorithms for Cox, DeepSurv, and Cox-Time can be naively implemented in practice, but not theoretically justifiable due to the form of the objective functions.}

\subsubsection{DE-constrained optimization}
 DE-constrained optimization has wide and important applications in various areas, such as optimal control, inverse problems, and shape optimization \citep{Antil2018}. One of the major contributions of this work is to recognize that the maximum likelihood estimation in survival analysis is essentially a DE-constrained optimization problem. Specifically, the maximum likelihood estimation (MLE) for the proposed SODEN can be rewritten as 
\begin{align}
	\max_{\theta} & \sum_{i=1}^N \Delta_i \log h(\Lambda_{x_i}(y_i), y_i; x_i, \theta) - \Lambda_{x_i}(y_i) \label{eq: log-mle} \\
	\text{subject to } & \Lambda'_{x_i}(t) = h(\Lambda_{x_i}(t), t; x_i, \theta) \nonumber \\
	& \Lambda_{x_i}(0)=0, ~i=1, \ldots, N \nonumber
\end{align}
where the constraint is a DE parameterized by $\theta$ and the objective contains the solution of the DE. Therefore, maximizing the likelihood function (\ref{general likelihood}) that contains the solution of the parameterized ODE can be viewed as an optimization problem with DE constraints as shown in (\ref{eq: log-mle})\footnote[2]{The optimization problem in (\ref{eq: log-mle}) belongs to a subclass of DE-constrained optimization problems, with the generic form of $\min_{\theta} J(\Lambda, \theta)$, subject to $g_1(\Lambda(t), \Lambda'(t), t;\theta)=0$ and $g_2(\Lambda(0); \theta)=0$.}. By bringing the strength of existing DE-constrained optimization techniques, we are able to develop novel numerical approaches for MLE in survival analysis without compromising the flexibility of models. There has been a rich literature on evaluating the gradient of the objective function in the DE-constrained optimization problem \citep{peto1972asymptotically, cao2003adjoint, alexe2009forward, gerdts2011optimal}. Among them, the adjoint sensitivity analysis is computationally efficient when evaluating the gradient of a scalar function with respect to large number of model parameters \citep{cao2003adjoint}. Therefore, we use the adjoint method to compute the gradient of (\ref{eq: log-mle}), whose detailed derivation is provided in Section~\ref{sub: adjoint}.

DE-constrained optimization has also found its applications in deep learning. \citet{chen2018neural} and \citet{dupont2019augmented} recently used ODEs parameterized with neural networks to model continuous-depth neural networks, normalizing flows, and time series, which lead to DE-constrained optimization problems. In this work, we share the merits of parameterizing the ODEs with neural networks but study a novel application of DE-constrained optimization in survival analysis. 


\section{The Proposed Approach}
\label{sec:approach}

\subsection{Survival Model through ODE Networks}
\label{model}
We consider the cumulative hazard function $\Lambda_{x}(\cdot)$ through an ODE (\ref{eq:ode}) with a fixed initial value. For readers' convenience, we repeat it below:
\begin{align*}
\left\{
\begin{array}{lr}
\Lambda'_x(t) = h(\Lambda_x(t), t; x, \theta) \\
\Lambda_x(0) =0
\end{array}
\right.,
\end{align*}
where the function $h$ determines the dynamic change of $\Lambda_{x}(\cdot)$: the derivative of cumulative hazard at time $t$ is determined by the current cumulative hazard $\Lambda_x(t)$, the current time $t$, and feature $x$ through the function $h$ parameterized by $\theta$. The initial value implies that the event always occurs after time $0$ since $S_x(0)=\exp(-\Lambda_x(0))=1$. Given an individual's feature vector $x$ and the parameter vector $\theta$, for any specific time point $t^*$, the cumulative hazard $\Lambda_x(t^*)$ can be obtained as the solution of the initial value problem (\ref{eq:ode}) at the time $t^*$, and the hazard rate can be obtained as $\lambda_x(t^*) = h(\Lambda_x(t^*), t^*; x, \theta)$. Therefore, the function $h$ fully determines the conditional distribution of the event time $T$ as shown in Section~\ref{prob}. 
The existence and uniqueness of the solution can be guaranteed if $h$ and its derivatives are Lipschitz continuous~\citep{Walter1998}. In this paper, we specify $h$ as a neural network and the above guarantees hold as long as the neural network has finite weights and Lipschitz non-linearities. In practice, we do not require the initial value problem (\ref{eq:ode}) to have a closed-form solution. We can obtain $\Lambda_x(t^*)$ numerically using any ODE solver given the derivative function $h$, initial value at $t_0=0$, evaluating time $t_1=t^*$, parameters $\theta$, and features~$x$, that is
\begin{equation}
\label{ODESolver}
    \Lambda_x(t^*) = \text{ODESolver}(h, \Lambda_{x}(0)=0,t_1 = t^*, x, \theta).
\end{equation}

We consider a general ODE form, where $h(\cdot, t; x, \theta)$ is a feed-forward neural network taking $\Lambda_x(t)$, $t$, and $x$ as inputs, and $\theta$ represents all parameters in the neural network. Specifically, the Softplus activation function~\citep{dugas2001incorporating} is used to constrain the output of the neural network, i.e. the hazard function, to be always positive.  We refer this general form as SODEN; note that SODEN is a flexible survival model as it does not make strong assumptions on the family of the underlying distribution or how features $x$ affect the event time. 

\paragraph{Remark 1.} Although there are other modeling alternatives that can uniquely characterize the event distribution as mentioned in Section~\ref{sec: prob func}, we choose to model the hazard in ODE~(\ref{eq:ode}) for three reasons. First, the hazard function has been widely used as the modeling target for summarizing survival data in the literature, due to its meaningful interpretation and informativeness about the underlying mechanism of events \citep[Chapter 2]{Klein2003}. Next, the hazard function is easier to model compared to the survival function, in the sense that it requires fewer constraints for the neural network structure under the ODE framework. For example, if we replace the cumulative hazard $\Lambda_x(t)$ with $S_x(t)$ in ODE (\ref{eq:ode}), we need to make sure the solution not only being monotonically decreasing in $t$ but also being within $[0, 1]$ for any $t\geq 0$, which poses additional constraints on the structure of the neural network $h$. Last but not least, the hazard function itself is of direct interest in many applications. For example, recent works in operational planning requires knowledge of the hazard rate of the waiting time until the customer abandons the queue \citep{doi:10.1287/mnsc.1090.1041,doi:10.1287/opre.1120.1069}.

\subsection{Model Learning}
\label{sec:learning}
We optimize SODEN by maximizing the likelihood function~(\ref{general likelihood}) given i.i.d. observations. The negative log-likelihood function of the $i^\text{th}$ observation can be written~as 
\begin{equation}
\label{nll}
    \mathcal{L}(\theta; D_i) \triangleq - \Delta_i \log h(\Lambda_{x_i}(y_i), y_i; x_i, \theta) + \Lambda_{x_i}(y_i),
\end{equation}
where $\Lambda_{x_i}(y_i)$, as given in (\ref{ODESolver}), also depends on parameters~$\theta$. Our goal is to minimize $\sum_{i=1}^N \mathcal{L}(\theta; D_i)$ with respect to~$\theta$.

For large-scale applications, we propose to use mini-batch SGD to optimize the criterion, where the gradient of $\mathcal{L}$ with respect to $\theta$ is calculated through the adjoint method \citep{pontryagin1962mathematical}. In comparison to naively applying the chain rule through all the operations used in computing the loss function, the adjoint method has the advantage of reducing memory usage and controlling numerical error explicitly in back-propagation. 

Next, we demonstrate how the gradients can be obtained. 
\subsubsection{Back-propagation through adjoint sensitivity analysis}
\label{sub: adjoint}
In the forward pass, we need to evaluate $\mathcal{L}(\theta; D_i)$ for each $i$ in a batch. While there might be no closed form for the solution of (\ref{eq:ode}), $\Lambda_{x_i}(y_i)$ can be numerically calculated using a black-box $\text{ODESolver}$ in (\ref{ODESolver}) and all other calculations are straightforward. In the backward pass, the only non-trivial part in the calculation of the gradients of $\mathcal{L}$ with respect to $\theta$ is back-propagation through the black-box $\text{ODESolver}$ in (\ref{ODESolver}). We compute it by solving another augmented ODE introduced by adjoint sensitivity analysis.
Specifically, let the adjoint  $a(t)$ satisfy $a'(t)= - \frac{\partial h}{\partial \Lambda} a(t)$ with $a(y_i)=1$, and then it follows that $\grad_{\theta} \Lambda_{x_i}(y_i) = \int_{0}^{y_i}a\frac{\partial h}{\partial \theta}\dev t$. Therefore, the gradient can be obtained by evaluating the following augmented ODE
\begin{equation}
\label{augode}
\left\{
\begin{array}{lr}
s'(t) = [h(\Lambda(t), t; x_i, \theta), -a(t)\frac{\partial h}{\partial \Lambda}, -a(t)\frac{\partial h}{\partial \theta}] \\
s(y_i) = [\Lambda_{x_i}(y_i), 1, \textbf{0}_{|\theta|}]
\end{array}
\right.,
\end{equation}
with $s(t) = [\Lambda(t), a(t), \bar{s}(t)]$ at $t=0$, i.e., $\grad_{\theta} \Lambda_{x_i}(y_i) = \bar{s}(0)$. 
Note that this approach does not need to access internal operations of ODE solvers used in the forward pass. Moreover, modern ODE solvers allow one to control the trade-off between the computing time and accuracy. 
Also note that a GPU-based implementation of back-propagation following the above rule is available in the \textit{torchdiffeq} library \citep{chen2018neural}. We provide the detailed derivation of (\ref{augode}) in Appendix~\ref{s: derivation of grad} for presentation integrity.

\subsubsection{Mini-batching with time-rescaling trick} 
We also provide a practical time-rescaling trick for mini-batching to better exploit the existing GPU-based implementation of ODE solvers. Concatenating ODEs of different observations in a mini-batch into a single combined ODE system is a useful trick for efficiently solving multiple ODEs on GPU. However, the existing GPU-based ODE solvers and the adjoint method in \citet{chen2018neural} require that all the individual ODEs share the same initial point $t_0$ and the evaluating point $t_1$ in the ODESolver (\ref{ODESolver}), which is unfortunately not the case in SODEN. For the $i^{th}$ observation in a mini-batch, the ODE (\ref{eq:ode}) in the forward pass needs to be evaluated at the corresponding observed time $t_1=y_i$.
To mitigate this discrepancy, we propose a time-rescaling trick that allows us to get the solution of individual ODEs at different time points by evaluating the combined ODE at only one time point. The key observation is that we can align the evaluating points of individual ODEs by variable transformation.
Let $H_i(t) = \Lambda_{x_i}(t\cdot y_i)$, for which the dynamics is determined by 
\begin{equation*}
\left\{
\begin{array}{lr}
H_i'(t) = h(H_i(t), ty_i; x_i, \theta)y_i \triangleq \tilde{h}(H_i(t), t;(x_i, y_i),\theta) \\
H_i(0) = \Lambda_{x_i}(0\cdot y_i) = 0
\end{array}
\right..
\end{equation*}
Since $H_i(1) = \Lambda_{x_i}(y_i)$ for all $i$, evaluating the combined ODE of all $H_i(s)$ at $s=1$ once will give us the values of $\Lambda_{x_i}(y_i)$ for all $i$. We therefore can take advantage of the existing GPU-based implementation for mini-batching by solving the combined ODE system of $H_i(s)$ with the time-rescaling trick\footnote{We note that some recently developed deep learning libraries (e.g., JAX~\citep{jax2018github}) could support mini-batching over complicated operations such as solving ODEs with different initial and evaluating time points without using the time-rescaling trick. However, the proposed rescaling trick provides an easy-to-implement extension for the \textit{torchdiffeq} library and potentially other frameworks.}.


\section{Simulation Study}
\label{sec:simulation}
In this section, we conduct a simulation study to illustrate that the proposed SODEN can fit well with data when the commonly used PH assumption does not hold.  For ease of visualization, we consider events generated from two groups where their survival functions cross each other, thus the PH assumption is violated. 
Further, we also show the advantage of SODEN as a continuous-time model rather than a discrete-time model. 

\subsection{Set-up}
We generate event times from the conditional distribution defined by the survival function $S_x(t) = e^{-2t}I(x=0) + e^{-2t^2}I(x=1)$,
where $x$ follows a Bernoulli distribution with probability $0.5$ and $I(\cdot)$ is the indicator function. The binary feature $x$ can be viewed as an indicator for two groups of individuals.
Note that the survival functions of the two groups, $S_0(t)$ and $S_1(t)$, cross at $t=1$, hence the PH assumption does not hold. The censoring times were uniformly sampled between $(0,2)$, which led to a censoring rate around 25\%.

We apply the proposed SODEN and investigate the predicted survival functions and hazard functions under $x=0$ and $x=1$ respectively. We also provide the results of DeepHit~\citep{lee2018deephit}, which is a discrete-time model without the PH assumption\footnote{See Appendix~\ref{s: discrete} for more details about this model.}, 
to further illustrate the advantage of the continuous nature of SODEN. We train both models on the same simulated data with sample size 10,000. The reported results are based on 10 independent trials.

\subsection{Results}

\begin{figure*}[t]
\centering
\begin{subfigure}{0.5\textwidth}
  \centering
  \includegraphics[width=0.83\linewidth]{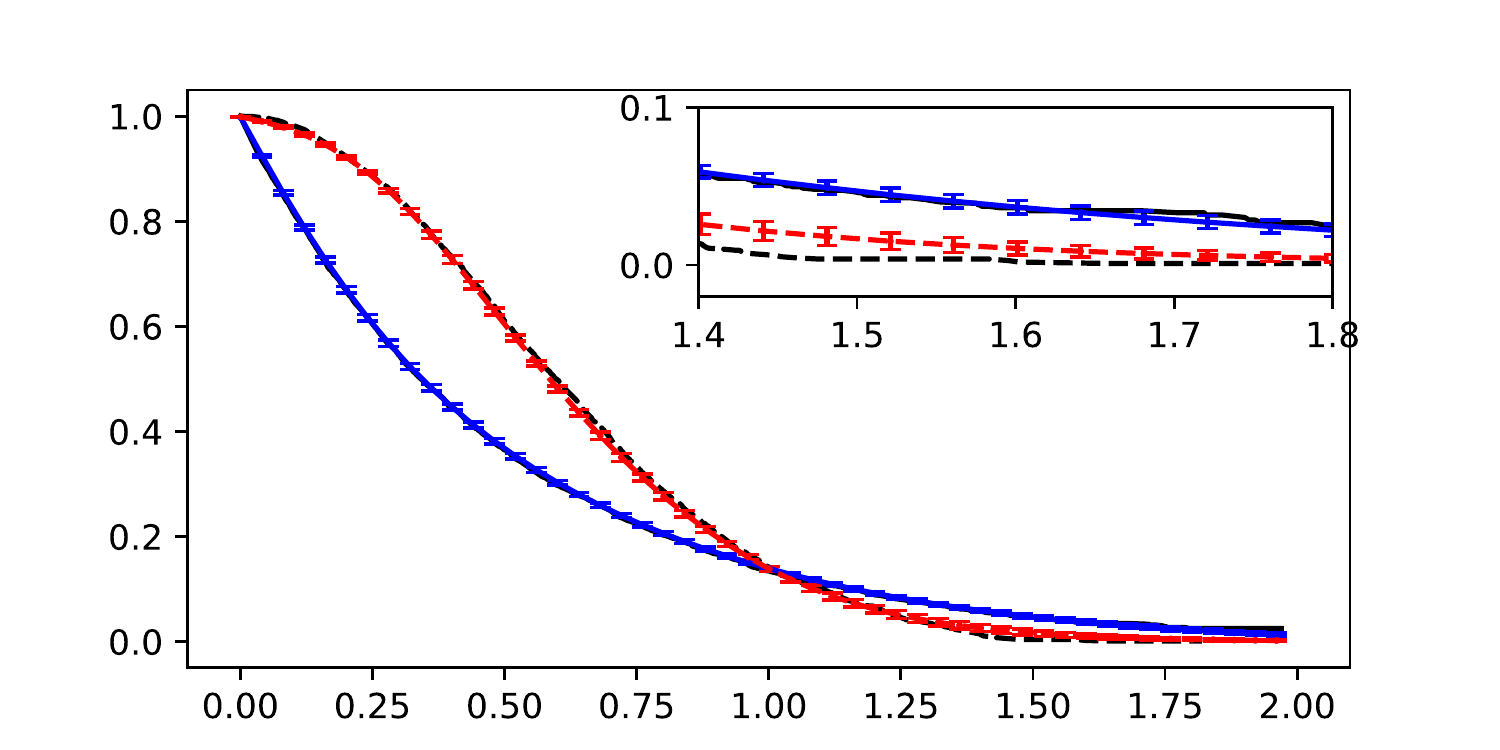}
\end{subfigure}%
\begin{subfigure}{0.5\textwidth}
  \centering
  \includegraphics[width=0.83\linewidth]{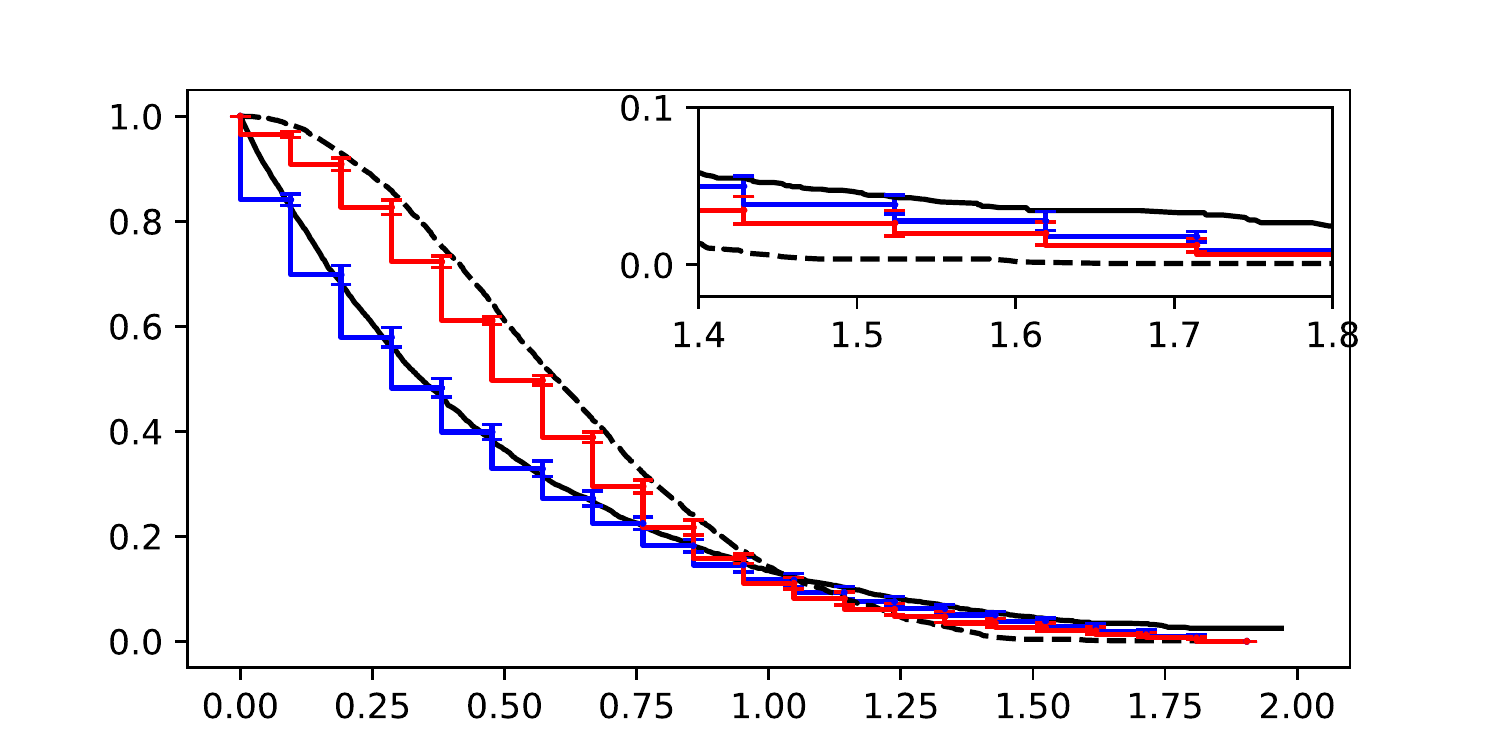}
\end{subfigure}
\begin{subfigure}{0.5\textwidth}
  \centering
 \includegraphics[width=0.83\linewidth]{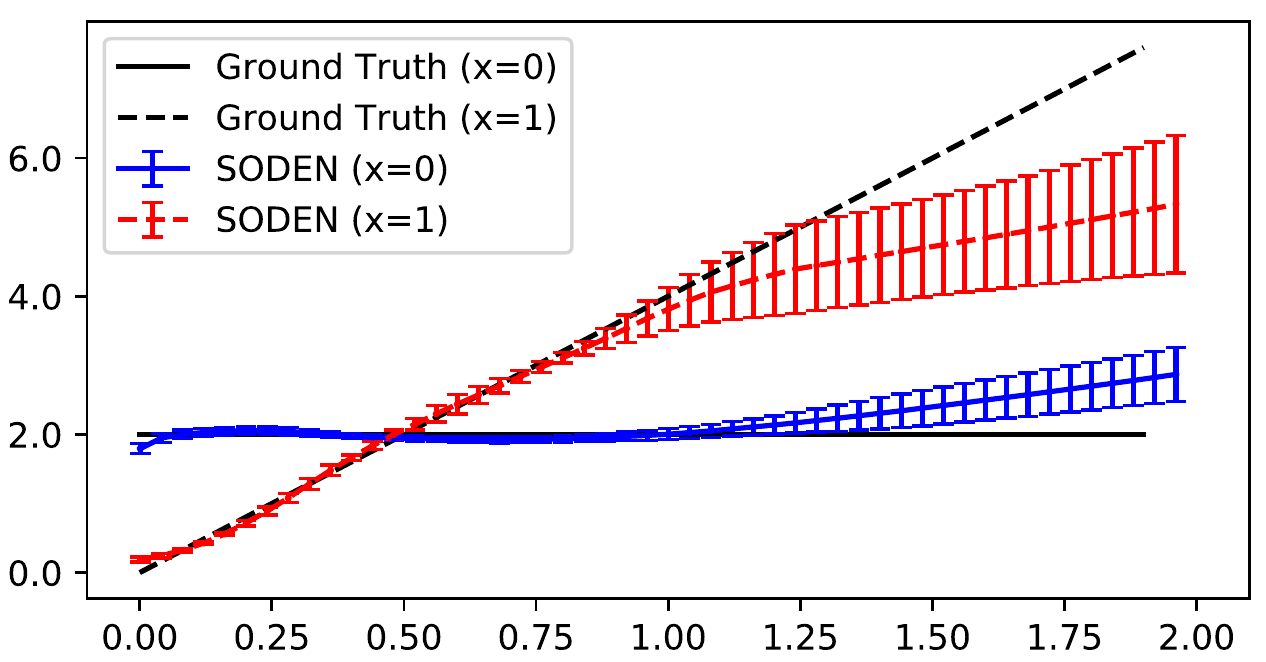}
\end{subfigure}%
\begin{subfigure}{0.5\textwidth}
  \centering
  \includegraphics[width=0.83\linewidth]{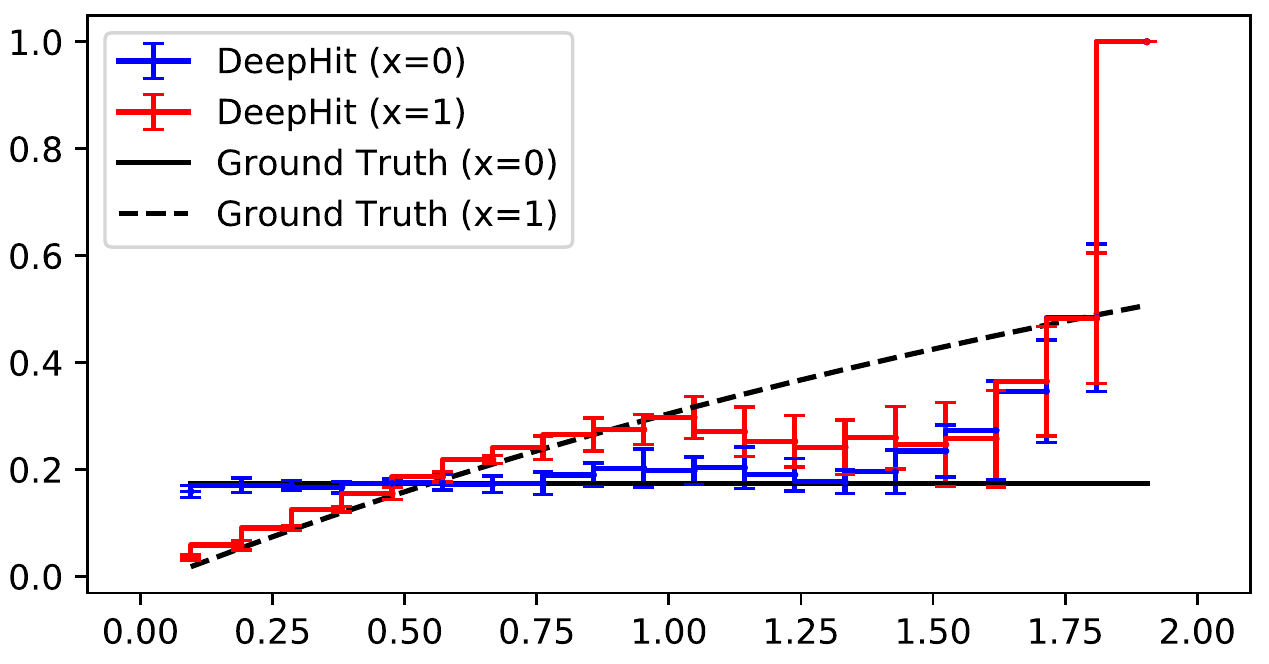}
\end{subfigure}
\caption{The survival functions (top row) and hazard functions (bottom row) of two groups, $x=0$ and $x=1$. The left column shows the results of SODEN, and the right column shows the results of DeepHit. In all figures, the results are the average of 10 independent trials and error bars indicate the standard deviation. The red curve indicates the predicted function for group $x=1$ and the blue curve indicates the predicted function for group $x=0$. The survival (Kaplan-Meier curves) and hazard functions corresponding to the data generating distribution for the two groups are shown in black curves (solid curves for group $x=0$ and dashed curves for group $x=1$). }
\label{fig:simulation}
\end{figure*}

The results of SODEN are shown in the left column of Figure~\ref{fig:simulation}. 
Note that the Kaplan-Meier (KM) estimate for each group can be considered a gold standard under our simulation setting, and we also plot them in Figure~\ref{fig:simulation} as the true survival functions corresponding to the data generating distribution. 
The predicted survival functions generally agree well with the true survival functions (the upper-left figure). The predicted survival functions of the two groups cross approximately at $t=1$, indicating SODEN can fit well with data not under the PH assumption. The lower-left figure shows that the predicted hazard functions of SODEN agree well with the true hazard functions when time is relatively small, but deviate from the true hazard functions as time increases. This is anticipated as there are few data points when $t$ is large and there are many more data points when $t$ is small. As a side note, while the estimate of the survival function looks better than that of the hazard function when $t$ is large, it is a visual artifact. As the survival function is monotonically decreasing and bounded between $0$ and $1$, the deviation (as indicated by the error bar) of the estimated survival function from the ground truth near the tail is visually diminished. Relatively, the estimate of the survival function actually becomes worse for larger time $t$.

The results of DeepHit are shown in the right column of Figure~\ref{fig:simulation}. Due to the discrete nature of the model, both the survival functions and the hazard functions predicted by DeepHit are step functions. While the predicted survival functions (the upper-right figure) fit well with the true survival functions when $t$ is small, the survival functions of the two groups are not well separated when $t$ is large. As for the hazard function (the lower-right figure), similarly, the predicted hazard functions fit well when $t$ is small but fluctuate wildly when $t$ is large.


\section{Real-world Examples}
\label{sec:real}
In this section, we demonstrate the effectiveness of SODEN by comparing it with five baseline models on three real-world datasets. We also conduct an ablation study to show the benefits of not making the PH assumption.

\subsection{Datasets}

We conduct experiments on the following three datasets: the Study to Understand Prognoses Preferences Outcomes and Risks of Treatment (\textbf{SUPPORT}), the Molecular Taxonomy of Breast Cancer International Consortium (\textbf{METABRIC})~\citep{katzman2018deepsurv}, and the Medical Information Mart for Intensive Care III (\textbf{MIMIC}) database~\citep{johnson2016mimic,goldberger2000physiobank}. 

SUPPORT and METABRIC are two common survival analysis benchmark datasets, which have been used in many previous works~\citep{katzman2018deepsurv,lee2018deephit,gensheimer2018simple,kvamme2019time}. We adopt the version pre-processed by \citet{katzman2018deepsurv} and refer readers there for more details. Despite their wide adoption in existing literature, we note that SUPPORT and METABRIC have relatively small sample sizes (8.8k for SUPPORT and 1.9k for METABRIC), which may not be ideal to evaluate deep survival analysis models. 

In this paper, we further build a novel large-scale survival analysis benchmark dataset from the publicly available MIMIC database. The MIMIC database provides deidentified clinical data of patients admitted to an Intensive Care Unit (ICU) stay. We take adult patients who are alive 24 hours after the first admission to ICU. 
The event of interest is defined as the mortality after admission. The event time is observed if there is a record of death in the database; otherwise, the censored time is defined as the last time of being discharged from the hospital. 
In MIMIC dataset, we extract 26 features based on the first 24-hour clinical data following \citet{purushotham2017benchmark}. In addition, to further evaluate deep learning models on applications with more complex data structure, we consider another feature set involving time series for the same group of patients, which is named as \textbf{MIMIC-SEQ} for differentiation. MIMIC-SEQ contains $5$ time-static features and $15$ time series features within the first $24$ hours after admission.  
Following the protocols described above, we are able to get a dataset with over 35k samples.

The detailed summary statistics of the three datasets are provided in Table~\ref{Table: datasets}. In all datasets, the categorical features are encoded as dummy variables and all the features are standardized. 

\begin{table*}
  \centering
  \begin{tabular}{lrcccccc}
    \toprule
    \multirow{2}{1.0cm}{Dataset} & \multirow{2}{0.6cm}{N} & \multirow{2}{0.3cm}{p} & \multirow{2}{1.5cm}{Censoring rate}  & \multicolumn{2}{c}{Censored time \small{(Yrs)}} & \multicolumn{2}{c}{Observed time \small{(Yrs)}} \\
    \cmidrule(r){5-8}
     & & & & Mean & Median & Mean & Median \\
    \midrule
    \small{MIMIC} & 35,304 & 26 & 61\% & 0.21  & 0.02 & 1.50 & 0.42 \\
    \small{(MIMIC-SEQ)} &  & \multirow{1}{1.3cm}{(5+15$\times$24)} & & & & & \\
    \small{SUPPORT} & 8,873 & 14 & 32\% & 2.90  & 2.51 & 0.56 & 0.16 \\
    \small{METABRIC} & 1,904 & 9 & 42\% & 0.44  & 0.43 & 0.27 & 0.24 \\
    \bottomrule
  \end{tabular}
   \caption{Summary statistics of three datasets. $N$ is the sample size and $p$ is the number of features. MIMIC-SEQ uses $5$ time-static features and $15$ time series features within the first $24$ hours after admission.}
  \label{Table: datasets}
\end{table*}

\subsection{Models for Comparison}
\label{sec:baselines}
We compare the proposed method with the classical linear \textbf{Cox} model and four state-of-the-art neural-network-based models:
\begin{itemize}
    \item \textbf{DeepSurv} is a PH model which replaces the linear feature combination in Cox with a neural network to improve feature extraction~\citep{katzman2018deepsurv}. 
    \item \textbf{Cox-Time} is a continuous-time model allowing non-PH, and is optimized by maximizing a modified partial-likelihood based loss function~\citep{kvamme2019time}. 
    \item \textbf{DeepHit} is a discrete-time survival model which estimates the probability mass at each pre-specified time interval, and is optimized by minimizing the linear combination of the negative log-likelihood and a differentiable surrogate ranking loss tailored for concordance index~\citep{lee2018deephit}.
    \item \textbf{Nnet-Survival} also models discrete-time distribution via estimating the conditional hazard probability at each time interval~\citep{gensheimer2018simple}.
\end{itemize}
Detailed model specifications and loss functions for the neural-network-based baselines can be found in Appendices~\ref{s: partial} and~\ref{s: discrete}. Note that on the MIMIC-SEQ dataset, we only compare neural-network-based models.

In Section~\ref{sec:simulation}, we have shown that the proposed model, because of its flexible parameterization, is able to fit well to the simulated data where the PH assumption does not hold. Here we further conduct an ablation study on real-world datasets to test the effect of the flexible parameterization. Specifically, we compare the general form of the proposed \textbf{SODEN}, with two of its degenerate variants, \textbf{SODEN-PH} and \textbf{SODEN-Cox}. SODEN-PH factorizes $h(\Lambda_x(t), t; x, \theta) = h_0(t;\theta)g(x;\theta)$ as a multiplication of two functions to satisfy the PH assumption, where both $h_0$ and $g$ are specified as neural networks. SODEN-Cox is a linear version of SODEN-PH where $g(x) = e^{x\beta}$. Notably, SODEN-Cox and SODEN-PH are designed to have similar representation power as Cox and DeepSurv respectively. 

\subsection{Evaluation Metrics}
\label{sec: eval}
Evaluating survival predictions needs to account for censoring. Here we describe several commonly used evaluation metrics~\citep{kvamme2019time,wang2019machine}.

\subsubsection{Time-dependent concordance index} Concordance index (C-index)~\citep{harrell1984regression} is a commonly used discriminative evaluation metric in survival analysis, and it measures the probability that, for a random pair of observations, the relative order of the two event times is consistent with that of the two predicted survival probabilities. The C-index was originally designed for models using the PH assumption, where the relative order of the predicted survival probabilities for two given individuals does not change with time. \citet{antolini2005time} further propose time-dependent C-index for models without PH assumption, where the relative order of the predicted survival probabilities may be different if evaluated at different time points. In addition, \citet{uno2011c} introduce inverse probability weights to the C-index such that it does not depend on the study-specific censoring distribution. Following \citet{antolini2005time} and \citet{uno2011c}, we adopt the inverse probability weighted time dependent C-index in our evaluation, which is given by
\[C^{td} = \frac{\sum_{i:\Delta_i=1}\sum_{j: y_i < y_j}I(\hat{S}_{x_i}(y_i) < \hat{S}_{x_j}(y_i))/ \hat{G}^2(y_i)}{\sum_{i:\Delta_i=1}\sum_{j: y_i < y_j}1/ \hat{G}^2(y_i)},\]
where $x_i$, $y_i$, and $\Delta_i$ are the features, observed time, and event indicator for individual $i$; $I(\cdot)$ is the indicator function; $\hat{S}_{x_i}(t)$ is the predicted survival function at time $t$ given $x_i$; and $\hat{G}(t)$ is the Kaplan-Meier estimator for the survival function of the censoring time, i.e. $\mathcal{P}(C>t)$. Under the independence assumption between the censoring time and the event time, $C^{td}$ converges to the discrimination measure $\mathcal{P}(S_{x_i}(T_i)<S_{x_j}(T_i)|T_i<T_j)$. 

In practice, the estimation of $\hat{G}(t)$ as well as the model predictions are relatively unstable for large $t$ due to limited number of observations, yet they lead to large inverse probability weights $1/\hat{G}(t)$. Following \citet{uno2011c}, we implement a truncated version of time-dependent C-index within a pre-specified time interval $(0,\tau)$, i.e., 
    \[C^{td}_\tau = \frac{\sum_{i:\Delta_i=1, y_i<\tau}\sum_{j: y_i < y_j}I(\hat{S}_{x_i}(y_i) < \hat{S}_{x_j}(y_i))/ \hat{G}^2(y_i)}{\sum_{i:\Delta_i=1, y_i<\tau}\sum_{j: y_i < y_j}1/ \hat{G}^2(y_i)}.\]
    We report results under various $\tau$ with $\hat{G}(\tau)=10^{-8}, 0.2$, and $0.4$. When $\hat{G}(\tau)=10^{-8}$, it is almost identical to the non-truncated version.  
Note that $C^{td}_{\tau}$=1 corresponds to a perfect ranking of predicted survival probabilities and $C^{td}_{\tau}$=0.5 corresponds to a random ordering.

\subsubsection{Integrated Brier score} 

For a binary classifier, the Brier score (BS) is defined as the mean square difference between the predicted probability and the ground-truth binary label. The metric BS can be decomposed into two components measuring calibration and discriminative performance respectively. Given similar discriminative performance, a lower BS indicates the closer the predicted survival probability $\hat{S}_x(t)$ is to the true probability of experiencing the event after time $t$. We refer well calibrated models to those with good probability estimates. \citet{graf1999assessment} generalized BS to take account for censoring in survival analysis. Specifically, the BS for survival analysis at time $t$ is defined as
\begin{align*}
	\text{BS}(t)& =\frac{1}{N}\sum_{i=1}^N \left\{\frac{(\hat{S}_{x_i}(t))^2I(y_i\leq t, \Delta_i=1)}{\hat{G}(y_i)}+\frac{(1-\hat{S}_{x_i}(t))^2I(y_i> t)}{\hat{G}(t)}\right\},
\end{align*}
where the notations are the same as $C^{td}$. As the predicted survival probability depends on the time point of evaluation, we use integrated BS (IBS) to measure the overall BS on a time interval: 
\[\text{IBS} = \frac{1}{t_{\text{max}}-t_{\text{min}}} \int_{t_{\text{min}}}^{t_{\text{max}}} \text{BS}(t) \dev t.\]
In practice, we choose the interval $[0,t_{\text{max}}]$ with various $t_{\text{max}}$ satisfying $\hat{G}(t_{\text{max}})=10^{-8}, 0.2$, and $0.4$, and compute this integral numerically by averaging over 100 grid points.  
The higher the IBS, the better the performance.

\subsubsection{Integrated binomial log-likelihood}
\citet{graf1999assessment} also generalized the binomial log-likelihood (BLL), which is a binary classification evaluation metric measuring both discrimination and calibration, to survival analysis in a similar way as BS. The BLL for survival analysis at time $t$ is defined as
\begin{align*}
	\text{BLL}(t)& =\frac{1}{N}\sum_{i=1}^N \left\{\frac{\log(1-\hat{S}_{x_i}(t))I(y_i\leq t, \Delta_i=1)}{\hat{G}(y_i)}+\frac{\log(\hat{S}_{x_i}(t))I(y_i> t)}{\hat{G}(t)}\right\},
\end{align*}
where the notations are the same as BS. We can also define the integrated BLL (IBLL) to measure the overall performance from $t_{\text{min}}$ to $t_{\text{max}}$, where
\[\text{IBLL} = \frac{1}{t_{\text{max}}-t_{\text{min}}} \int_{t_{\text{min}}}^{t_{\text{max}}} \text{BLL}(t) \dev t.\]
The higher the IBLL, the better the performance. Note that the IBS takes the squared error in the loss, i.e., $\text{error}^2$, while the negative IBLL accounts for error with scale $-\log(1-\text{error})$. Thus, in general, IBLL has larger magnitude than IBS and penalizes more for larger error.

\subsubsection{Negative log-likelihood} Negative log-likelihood (NLL) corresponds to $\mathcal{L}(\theta; D_i)$ in (\ref{nll}) and predictive NLL on held out data measures the goodness-of-fit of the model to the observed data. However, NLL is only applicable to models that provide likelihood, and it is not comparable between discrete-time and continuous-time models due to the difference in the likelihood definition. We use NLL to compare three variants of SODEN in the ablation study. The lower the NLL, the better the performance.

\subsection{Experimental Setup}

We randomly split each dataset into training, validation and testing sets with a ratio of 3:1:1. To make the evaluation more reliable, we take 5 independent random splits for MIMIC(-SEQ), 10 independent random splits for SUPPORT and METABRIC as their sizes are relatively small. For each split, we train the Cox model on the combination of training and validation sets. For neural-network-based models, we train each model on the training set, and apply early-stopping using the loss on the validation set with patience 10. The hyper-parameters of each model are tuned within each split through 100 independent trials using random search. We select the optimal hyper-parameter setting with the best score on the validation set. For continuous-time models, DeepSurv, Cox-Time, and SODEN, the validation score is set as the loss. For discrete-time models, DeepHit and Nnet-Survival, the loss functions (i.e., NLLs) across different pre-specified time intervals are not comparable so the validation score is set as $C^{td}$ as was done in \citet{kvamme2019time}. 

For all neural networks, we use multilayer perceptrons (MLP) with ReLU activation in all layers except for the output layer. For SODEN, Softplus is used to constrain the output to be always positive; for DeepHit and Nnet-Survival, Softmax and Sigmoid are used respectively to return PMF and discrete hazard probability. 
For the MIMIC-SEQ dataset, we incorporate a one-layer Gated Recurrent Units (GRU) encoder into the model architecture of each deep survival model to learn feature representation from sequence data.  
We use the RMSProp~\citep{tieleman2012lecture} optimizer and tune batch size, learning rate, weight decay, momentum, the number of layers, and the number of neurons in each layer. The search ranges for the aforementioned hyper-parameters are shared across all neural-network-based models on each dataset. Additionally, we tune batch normalization and dropout for all neural-network-based baseline models. For DeepHit and Nnet-Survival, we tune the number of pre-specified time intervals. We also smooth the predicted survival function by interpolation, which is an important post-processing step to improve the performance of these discrete-time models. The tuning ranges of hyper-parameters are listed in Appendix~\ref{s: hyper}.

\subsection{Results}
\begin{table}[!t]
\centering
\begin{tabular}{ll|lll}
\toprule
$\mathcal{P}(C>\tau)$ & Model &              $C^{td}_\tau$ ($\uparrow$) &                        IBLL ($\uparrow$) &                      IBS ($\downarrow$) \\
\midrule
$10^{-8}$ & DeepSurv &              $0.685 \pm \small{.002}$ &               $-0.335 \pm \small{.003}$ &      $\textbf{0.103} \pm \small{.001}$ \\
             & Cox-Time &              $0.681 \pm \small{.002}$ &               $-0.332 \pm \small{.003}$ &      $\textbf{0.103} \pm \small{.001}$ \\
             & Nnet-Survival &             $0.679 \pm \small{.003*}$ &  $\underline{-0.331} \pm \small{.003*}$ &               $0.104 \pm \small{.001}$ \\
             & DeepHit &     $\textbf{0.688} \pm \small{.002}$ &              $-0.336 \pm \small{.005*}$ &              $0.106 \pm \small{.001*}$ \\
             & SODEN (ours) &  $\underline{0.687} \pm \small{.002}$ &      $\textbf{-0.328} \pm \small{.004}$ &      $\textbf{0.103} \pm \small{.001}$ \\
\midrule
$0.2$ & DeepSurv &              $0.685 \pm \small{.002}$ &               $-0.400 \pm \small{.013}$ &      $\textbf{0.124} \pm \small{.003}$ \\
             & Cox-Time &              $0.681 \pm \small{.002}$ &               $-0.397 \pm \small{.011}$ &   $\underline{0.125} \pm \small{.003}$ \\
             & Nnet-Survival &             $0.679 \pm \small{.003*}$ &  $\underline{-0.396} \pm \small{.011*}$ &               $0.126 \pm \small{.003}$ \\
             & DeepHit &     $\textbf{0.688} \pm \small{.002}$ &              $-0.402 \pm \small{.012*}$ &              $0.128 \pm \small{.004*}$ \\
             & SODEN (ours) &  $\underline{0.687} \pm \small{.002}$ &      $\textbf{-0.391} \pm \small{.011}$ &   $\underline{0.125} \pm \small{.004}$ \\
\midrule
$0.4$ & DeepSurv &             $0.740 \pm \small{.002*}$ &              $-0.386 \pm \small{.013*}$ &              $0.121 \pm \small{.005*}$ \\
             & Cox-Time &             $0.744 \pm \small{.003*}$ &              $-0.382 \pm \small{.014*}$ &  $\underline{0.120} \pm \small{.005*}$ \\
             & Nnet-Survival &             $0.737 \pm \small{.004*}$ &               $-0.391 \pm \small{.015}$ &               $0.123 \pm \small{.006}$ \\
             & DeepHit &     $\textbf{0.752} \pm \small{.003}$ &   $\underline{-0.381} \pm \small{.014}$ &  $\underline{0.120} \pm \small{.005*}$ \\
             & SODEN (ours) &     $\textbf{0.752} \pm \small{.003}$ &      $\textbf{-0.374} \pm \small{.013}$ &      $\textbf{0.118} \pm \small{.005}$ \\
\bottomrule
\end{tabular}
\caption{Comparison of time dependent concordance index ($C^{td}_\tau$), integrated binomial log-likelihood (IBLL), integrated brier score (IBS) on MIMIC-SEQ. The \textbf{bold} and \underline{underline} markers denote the best and the second best performance respectively. The ($\pm$)~error bar denotes the standard error of the mean. The asterisk (*) after a baseline model performance indicates a significant (either positive or negative) difference between that baseline model and the proposed SODEN, under pairwise t-test with p-value $<0.05$.} 
\label{tab:MIMIC-SEQ}
\end{table}

\begin{table}[!t]
\centering
\begin{tabular}{ll|lll}
\toprule
$\mathcal{P}(C>\tau)$ & Model &              $C^{td}_\tau$ ($\uparrow$) &                        IBLL ($\uparrow$) &                      IBS ($\downarrow$) \\
\midrule
$10^{-8}$ & Cox &             $0.660 \pm \small{.001*}$ &              $-0.335 \pm \small{.003*}$ &              $0.105 \pm \small{.001*}$ \\
             & DeepSurv &              $0.683 \pm \small{.001}$ &              $-0.326 \pm \small{.005*}$ &  $\underline{0.101} \pm \small{.001*}$ \\
             & Cox-Time &              $0.680 \pm \small{.001}$ &              $-0.326 \pm \small{.003*}$ &  $\underline{0.101} \pm \small{.001*}$ \\
             & Nnet-Survival &              $0.681 \pm \small{.001}$ &   $\underline{-0.321} \pm \small{.002}$ &   $\underline{0.101} \pm \small{.001}$ \\
             & DeepHit &     $\textbf{0.685} \pm \small{.002}$ &              $-0.327 \pm \small{.003*}$ &              $0.102 \pm \small{.001*}$ \\
             & SODEN (ours) &  $\underline{0.684} \pm \small{.002}$ &      $\textbf{-0.319} \pm \small{.003}$ &      $\textbf{0.100} \pm \small{.001}$ \\
\midrule
$0.2$ & Cox &             $0.660 \pm \small{.001*}$ &              $-0.413 \pm \small{.007*}$ &              $0.132 \pm \small{.003*}$ \\
             & DeepSurv &              $0.683 \pm \small{.001}$ &              $-0.402 \pm \small{.006*}$ &  $\underline{0.127} \pm \small{.002*}$ \\
             & Cox-Time &              $0.680 \pm \small{.001}$ &              $-0.404 \pm \small{.007*}$ &              $0.128 \pm \small{.002*}$ \\
             & Nnet-Survival &              $0.682 \pm \small{.001}$ &   $\underline{-0.398} \pm \small{.007}$ &   $\underline{0.127} \pm \small{.003}$ \\
             & DeepHit &     $\textbf{0.685} \pm \small{.002}$ &              $-0.404 \pm \small{.008*}$ &              $0.128 \pm \small{.002*}$ \\
             & SODEN (ours) &  $\underline{0.684} \pm \small{.002}$ &      $\textbf{-0.395} \pm \small{.006}$ &      $\textbf{0.126} \pm \small{.002}$ \\
\midrule
$0.4$ & Cox &             $0.706 \pm \small{.003*}$ &              $-0.399 \pm \small{.018*}$ &              $0.124 \pm \small{.007*}$ \\
             & DeepSurv &             $0.739 \pm \small{.003*}$ &               $-0.387 \pm \small{.016}$ &  $\underline{0.120} \pm \small{.007*}$ \\
             & Cox-Time &             $0.737 \pm \small{.003*}$ &              $-0.387 \pm \small{.020*}$ &  $\underline{0.120} \pm \small{.007*}$ \\
             & Nnet-Survival &              $0.741 \pm \small{.005}$ &  $\underline{-0.386} \pm \small{.019*}$ &  $\underline{0.120} \pm \small{.007*}$ \\
             & DeepHit &     $\textbf{0.747} \pm \small{.004}$ &               $-0.404 \pm \small{.023}$ &               $0.128 \pm \small{.009}$ \\
             & SODEN (ours) &  $\underline{0.746} \pm \small{.003}$ &      $\textbf{-0.379} \pm \small{.019}$ &      $\textbf{0.118} \pm \small{.007}$ \\
\bottomrule
\end{tabular}
\caption{Comparison of performance on MIMIC. The notations share the same definitions as in Table~\ref{tab:MIMIC-SEQ}.}
\label{tab:MIMIC}
\end{table}

\begin{table}[!t]
\centering
\begin{tabular}{ll|lll}
\toprule
$\mathcal{P}(C>\tau)$ & Model &              $C^{td}_\tau$ ($\uparrow$) &                       IBLL ($\uparrow$) &                     IBS ($\downarrow$) \\
\midrule
$10^{-8}$ & Cox &             $0.596 \pm \small{.002*}$ &             $-0.568 \pm \small{.001*}$ &             $0.194 \pm \small{.001*}$ \\
             & DeepSurv &             $0.609 \pm \small{.003*}$ &    $\textbf{-0.559} \pm \small{.002*}$ &    $\textbf{0.190} \pm \small{.001*}$ \\
             & Cox-Time &             $0.607 \pm \small{.004*}$ &              $-0.565 \pm \small{.002}$ &  $\underline{0.191} \pm \small{.001}$ \\
             & Nnet-Survival &              $0.624 \pm \small{.003}$ &              $-0.570 \pm \small{.004}$ &             $0.193 \pm \small{.001*}$ \\
             & DeepHit &     $\textbf{0.631} \pm \small{.003}$ &             $-0.583 \pm \small{.006*}$ &             $0.197 \pm \small{.001*}$ \\
             & SODEN (ours) &  $\underline{0.627} \pm \small{.003}$ &  $\underline{-0.563} \pm \small{.002}$ &  $\underline{0.191} \pm \small{.001}$ \\
\midrule
$0.2$ & Cox &             $0.596 \pm \small{.002*}$ &             $-0.585 \pm \small{.001*}$ &             $0.201 \pm \small{.000*}$ \\
             & DeepSurv &             $0.609 \pm \small{.003*}$ &     $\textbf{-0.577} \pm \small{.002}$ &     $\textbf{0.197} \pm \small{.001}$ \\
             & Cox-Time &             $0.606 \pm \small{.004*}$ &              $-0.583 \pm \small{.002}$ &              $0.199 \pm \small{.001}$ \\
             & Nnet-Survival &              $0.623 \pm \small{.003}$ &              $-0.586 \pm \small{.003}$ &             $0.201 \pm \small{.001*}$ \\
             & DeepHit &     $\textbf{0.630} \pm \small{.003}$ &             $-0.601 \pm \small{.006*}$ &             $0.205 \pm \small{.002*}$ \\
             & SODEN (ours) &  $\underline{0.627} \pm \small{.003}$ &  $\underline{-0.579} \pm \small{.002}$ &  $\underline{0.198} \pm \small{.001}$ \\
\midrule
$0.4$ & Cox &             $0.595 \pm \small{.002*}$ &             $-0.602 \pm \small{.001*}$ &             $0.208 \pm \small{.001*}$ \\
             & DeepSurv &             $0.608 \pm \small{.002*}$ &     $\textbf{-0.595} \pm \small{.002}$ &     $\textbf{0.205} \pm \small{.001}$ \\
             & Cox-Time &             $0.605 \pm \small{.004*}$ &              $-0.601 \pm \small{.002}$ &              $0.207 \pm \small{.001}$ \\
             & Nnet-Survival &              $0.623 \pm \small{.003}$ &              $-0.602 \pm \small{.003}$ &             $0.208 \pm \small{.001*}$ \\
             & DeepHit &     $\textbf{0.630} \pm \small{.003}$ &             $-0.619 \pm \small{.007*}$ &             $0.212 \pm \small{.002*}$ \\
             & SODEN (ours) &  $\underline{0.626} \pm \small{.003}$ &  $\underline{-0.597} \pm \small{.002}$ &     $\textbf{0.205} \pm \small{.001}$ \\
\bottomrule
\end{tabular}
\caption{Comparison of performance on SUPPORT. The notations share the same definitions as in Table~\ref{tab:MIMIC-SEQ}.}
\label{tab:SUPPORT}
\end{table}

\begin{table}[!t]
\centering
\begin{tabular}{ll|lll}
\toprule
$\mathcal{P}(C>\tau)$ & Model &              $C^{td}_{\tau}$ ($\uparrow$) &                        $\text{IBLL}_{\tau}$ ($\uparrow$) &                     $\text{IBS}_{\tau}$ ($\downarrow$) \\
\midrule
$10^{-8}$ & Cox &             $0.644 \pm \small{.006*}$ &  $\underline{-0.508} \pm \small{.009*}$ &  $\underline{0.169} \pm \small{.002}$ \\
             & DeepSurv &             $0.635 \pm \small{.007*}$ &              $-0.517 \pm \small{.011*}$ &             $0.171 \pm \small{.003*}$ \\
             & Cox-Time &             $0.648 \pm \small{.007*}$ &              $-0.511 \pm \small{.009*}$ &             $0.172 \pm \small{.003*}$ \\
             & Nnet-Survival &  $\underline{0.666} \pm \small{.005}$ &               $-0.510 \pm \small{.007}$ &             $0.171 \pm \small{.002*}$ \\
             & DeepHit &    $\textbf{0.674} \pm \small{.006*}$ &              $-0.514 \pm \small{.004*}$ &             $0.174 \pm \small{.002*}$ \\
             & SODEN (ours) &              $0.661 \pm \small{.005}$ &      $\textbf{-0.498} \pm \small{.008}$ &     $\textbf{0.167} \pm \small{.003}$ \\
\midrule
$0.2$ & Cox &             $0.639 \pm \small{.006*}$ &   $\underline{-0.521} \pm \small{.006}$ &  $\underline{0.176} \pm \small{.002}$ \\
             & DeepSurv &             $0.635 \pm \small{.006*}$ &              $-0.530 \pm \small{.005*}$ &             $0.179 \pm \small{.002*}$ \\
             & Cox-Time &             $0.647 \pm \small{.005*}$ &              $-0.531 \pm \small{.007*}$ &             $0.179 \pm \small{.002*}$ \\
             & Nnet-Survival &  $\underline{0.662} \pm \small{.004}$ &               $-0.523 \pm \small{.003}$ &              $0.177 \pm \small{.001}$ \\
             & DeepHit &    $\textbf{0.671} \pm \small{.004*}$ &              $-0.533 \pm \small{.003*}$ &             $0.182 \pm \small{.001*}$ \\
             & SODEN (ours) &              $0.659 \pm \small{.003}$ &      $\textbf{-0.516} \pm \small{.005}$ &     $\textbf{0.174} \pm \small{.002}$ \\
\midrule
$0.4$ & Cox &             $0.637 \pm \small{.006*}$ &               $-0.521 \pm \small{.006}$ &  $\underline{0.175} \pm \small{.002}$ \\
             & DeepSurv &             $0.635 \pm \small{.006*}$ &              $-0.526 \pm \small{.005*}$ &             $0.178 \pm \small{.002*}$ \\
             & Cox-Time &             $0.644 \pm \small{.005*}$ &              $-0.526 \pm \small{.006*}$ &             $0.178 \pm \small{.002*}$ \\
             & Nnet-Survival &  $\underline{0.660} \pm \small{.003}$ &   $\underline{-0.519} \pm \small{.003}$ &              $0.176 \pm \small{.001}$ \\
             & DeepHit &    $\textbf{0.668} \pm \small{.004*}$ &              $-0.528 \pm \small{.003*}$ &             $0.180 \pm \small{.001*}$ \\
             & SODEN (ours) &              $0.658 \pm \small{.004}$ &      $\textbf{-0.513} \pm \small{.005}$ &     $\textbf{0.173} \pm \small{.002}$ \\
\bottomrule
\end{tabular}
\caption{Comparison of performance on METABRIC. The notations share the same definitions as in Table~\ref{tab:MIMIC-SEQ}.}
\label{tab:METABRIC}
\end{table}

\subsubsection{Discriminative and calibration performance}
The comparison of model performances on MIMIC-SEQ, MIMIC, SUPPORT, and METABRIC are respectively reported in Tables~\ref{tab:MIMIC-SEQ} to~\ref{tab:METABRIC}.

We first consider the C-index metric, which measures the discriminative performance. We observe that the proposed SODEN outperforms other continuous-time models (Cox, DeepSurv, and Cox-Time). The differences in C-index are significant on all datasets, except for those with large $\tau$ on MIMIC-SEQ and MIMIC. The gain of SODEN against DeepSurv and Cox-Time demonstrates the benefits of not making the PH assumption and having a principled likelihood objective. We also observe that all neural network models significantly outperform the Cox model in almost all cases. 

For discrete-time models, Nnet-Survival and DeepHit show strong discriminative performance on the C-index metric compared to continuous-time models in general. This is not surprising due to the facts that 1) similar as SODEN, the discrete-time models do not make strong structural assumptions; 2) the discrete-time models are tuned with C-index as the validation metric, and DeepHit has an additional ranking loss tailored for C-index. However, we find their advantage diminishes on MIMIC-SEQ and MIMIC, where the data size is much larger. We suspect the information loss due to discretizing the event time becomes more severe as the data size grows, and will eventually turn to the discriminative performance bottleneck.

We then consider the IBLL and IBS metrics, which measure a combination of the discriminative performance and the calibration performance. Overall, most models are similarly well-calibrated. However, DeepHit is obviously less calibrated than most other models, given it has the worst IBLL and IBS and the best C-index metric in most settings. This may be due to the surrogate ranking loss used in DeepHit.

In summary, the proposed SODEN demonstrates significantly better discriminative performance than all continuous-time baseline methods on all datasets. On the larger datasets (MIMIC-SEQ and MIMIC), SODEN achieves better or similar C-index metric compared to the discrete models. The superior discriminative performance of DeepHit comes at the price of the inferior calibration performance.

Finally, we remark that the event time and censoring time in MIMIC both have heavily right-skewed distributions, as indicated by the large discrepancy between their mean and median in Table~\ref{Table: datasets}. On MIMIC and MIMIC-SEQ, including more testing data near the tail in evaluation ($\hat{G}(\tau)=10^{-8}$ or $0.2$\footnote{On MIMIC and MIMIC-SEQ, both $\hat{G}(\tau)=10^{-8}$ and $\hat{G}(\tau)=0.2$ have a tiny number of samples being excluded due to the right-skewness of the censoring distribution, and thus are close to the non-truncated version $C^{td}$.}) gives a worse $C^{td}_{\tau}$ compared to including less tail samples ($\hat{G}(\tau)=0.4$). This is because models tend to have poor prediction performance near the tail due to limited number of observations, yet these tail samples get large inverse probability weights. This also explains why the differences in $C^{td}_{\tau}$ among different models are less significant when including more tail samples.

\begin{table}[!t]
\centering
\begin{tabular}{ll|lll}
\toprule
Dataset & Metric ($\mathcal{P}(C>\tau)$)&                               SODEN &                                SODEN-PH &                               SODEN-Cox \\
\midrule
MIMIC-SEQ & NLL &  $\textbf{0.489} \pm \small{.072}$ &  $\underline{0.520} \pm \small{.069*}$ &                                     N/A \\
         & $C^{td}$ ($10^{-8}$) &  $\textbf{0.687} \pm \small{.002}$ &  $\underline{0.682} \pm \small{.001*}$ &                                     N/A \\
         & $C^{td}$ ($0.2$) &  $\textbf{0.687} \pm \small{.002}$ &  $\underline{0.683} \pm \small{.001*}$ &                                     N/A \\
         & $C^{td}$ ($0.4$) &  $\textbf{0.752} \pm \small{.003}$ &  $\underline{0.739} \pm \small{.005*}$ &                                     N/A \\
\midrule
MIMIC & NLL &  $\textbf{0.411} \pm \small{.007}$ &  $\underline{0.436} \pm \small{.007*}$ &              $0.450 \pm \small{.006*}$ \\
         & $C^{td}$ ($10^{-8}$) &  $\textbf{0.684} \pm \small{.002}$ &  $\underline{0.679} \pm \small{.002*}$ &              $0.659 \pm \small{.001*}$ \\
         & $C^{td}$ ($0.2$) &  $\textbf{0.684} \pm \small{.002}$ &  $\underline{0.679} \pm \small{.002*}$ &              $0.659 \pm \small{.001*}$ \\
         & $C^{td}$ ($0.4$) &  $\textbf{0.746} \pm \small{.003}$ &  $\underline{0.734} \pm \small{.003*}$ &              $0.706 \pm \small{.003*}$ \\
\midrule
SUPPORT & NLL &  $\textbf{0.676} \pm \small{.008}$ &  $\underline{0.702} \pm \small{.008*}$ &              $0.761 \pm \small{.022*}$ \\
         & $C^{td}$ ($10^{-8}$) &  $\textbf{0.627} \pm \small{.003}$ &  $\underline{0.608} \pm \small{.003*}$ &              $0.591 \pm \small{.003*}$ \\
         & $C^{td}$ ($0.2$) &  $\textbf{0.627} \pm \small{.003}$ &  $\underline{0.608} \pm \small{.002*}$ &              $0.590 \pm \small{.004*}$ \\
         & $C^{td}$ ($0.4$) &  $\textbf{0.626} \pm \small{.003}$ &  $\underline{0.607} \pm \small{.002*}$ &              $0.589 \pm \small{.004*}$ \\
\midrule
METABRIC & NLL &  $\textbf{0.149} \pm \small{.015}$ &              $0.176 \pm \small{.013*}$ &  $\underline{0.167} \pm \small{.010*}$ \\
         & $C^{td}$ ($10^{-8}$) &  $\textbf{0.661} \pm \small{.005}$ &              $0.640 \pm \small{.005*}$ &  $\underline{0.642} \pm \small{.006*}$ \\
         & $C^{td}$ ($0.2$) &  $\textbf{0.659} \pm \small{.003}$ &  $\underline{0.639} \pm \small{.004*}$ &              $0.638 \pm \small{.005*}$ \\
         & $C^{td}$ ($0.4$) &  $\textbf{0.658} \pm \small{.004}$ &  $\underline{0.639} \pm \small{.005*}$ &              $0.636 \pm \small{.006*}$ \\
\bottomrule
\end{tabular}
\caption{Comparison of negative log-likelihood (NLL) and time dependent concordance index ($C^{td}_\tau$) between SODEN and its degenerate variants, SODEN-Cox and SODEN-PH, for ablation study. The \textbf{bold}, \underline{underline}, and ($\pm$)~error bar share the same definitions as in Table~\ref{tab:MIMIC-SEQ}. The asterisk (*) indicates a significant difference between the proposed SODEN and its degenerate variants, under pairwise t-test with p-value $<0.05$.}
\label{tab:ablation}
\end{table}

\begin{figure}[!t]
 \begin{center}
 \includegraphics[width=0.67\linewidth]{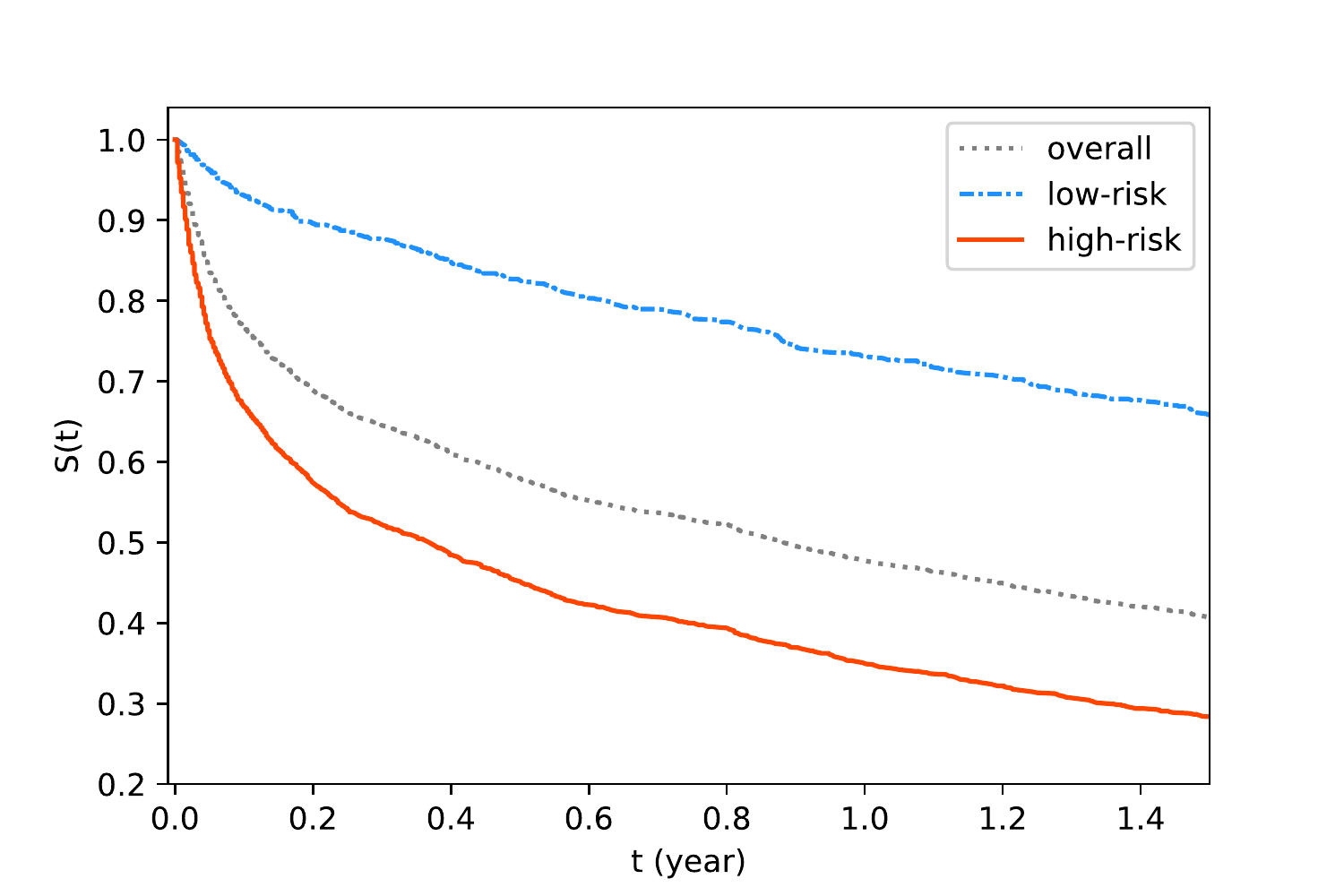}
 \end{center}
\caption{Kaplan-Meier curves of high/low-risk groups for SODEN on MIMIC.}
\label{fig:KM curve}
\end{figure}

\subsubsection{Ablation study}

While the trend over Cox, DeepSurv, and SODEN has supported our conjecture that flexible parameterization by introducing non-linearity and not making the PH assumption is important for practical survival analysis on modern datasets, we further verify this conjecture by the ablation study with SODEN-PH and SODEN-Cox (see Table~\ref{tab:ablation}). 

First, we observe that the relative differences in the C-index metric among SODEN-Cox, SODEN-PH, and SODEN are similar as those among Cox, DeepSurv, and SODEN. In fact, we can see that the $C^{td}_{\tau}$'s of SODEN-Cox and SODEN-PH in Table~\ref{tab:ablation} are respectively similar with those of their partial-likelihood counterparts Cox and DeepSurv in Tables~\ref{tab:MIMIC-SEQ} to~\ref{tab:METABRIC}. This observation implies that 1) neural networks can approximate the baseline hazard function as well as the non-parametric Breslow's estimator~\citep{breslow1972discussion}; 2) maximizing the likelihood function with numerical approximation approaches, where SGD based algorithms can be naturally applied, can perform as well as maximizing the partial likelihood for PH models. 

Second, SODEN outperforms SODEN-PH and SODEN-Cox in terms of NLL by a large margin. The major difference between SODEN-PH and SODEN is that the former is restricted by the PH assumption while the latter is not. The comparison of NLL between SODEN-PH and SODEN provides a strong evidence that the PH assumption may not hold on these datasets. Further, SODEN-Cox often being the worst verifies again that both non-linearity and the flexibility of non-PH models matter.

\subsubsection{Risk discriminating visualization} 
We further provide visualization of risk discrimination.  We show the Kaplan-Meier curves~\citep{kaplan1958nonparametric} of high-risk and low-risk groups identified by SODEN on the MIMIC dataset. We first obtain the predicted survival probability for each individual at the median of all observed survival times in the test set. We then split the test set into high-risk and low-risk groups evenly based on their predicted survival probabilities. 
The Kaplan-Meier curves for the high-risk group, the low-risk group, and the entire test set are shown in Figure~\ref{fig:KM curve}. The difference between high-risk and low-risk groups is statistically significant where the p-value of the log rank test \citep{peto1972asymptotically} is smaller than 0.001.


\section{Conclusion}
\label{sec:conclusion}
In this paper, we have proposed a survival model through ordinary differential equation networks. It can model a broad range family of continuous event time distributions without strong structural assumptions and can obtain powerful feature representations using neural networks. Moreover, we have tackled the challenge of evaluating the likelihood of survival models and the gradients with respect to model parameters by an efficient numerical approach. The algorithm scales well by allowing direct use of mini-batch SGD. We have also demonstrated the effectiveness of the proposed method on both simulation studies and real-world data examples. 


\acks{We would like to thank Yuanhao Liu and Chenkai Sun for running part of the baseline experiments; Yuekai Sun, Xuefei Zhang, and Debarghya Mukherjee for their feedback on the draft.}


\vskip 0.2in
\bibliography{reference}

\newpage
\appendix
\section{Derivation of Gradients through Adjoint Sensitivity Analysis}
\label{s: derivation of grad}
We rewrite $\Lambda_{x_i}(y_i)$ as the objective function $G(\Lambda, \theta)=\int^{y_i}_0 h(\Lambda(t), t, x_i;\theta)\dev t$ 
with the following DE constraint
\begin{equation}
\label{constraint}
\left\{
\begin{array}{lr}
\Lambda'(t) = h(\Lambda(t), t; x_i, \theta) \\
\Lambda(0) =0
\end{array}
\right.,
\end{equation}
where we simplify the notation $\Lambda_{x_i}$ as $\Lambda$. Now we wish to calculate the gradient of $G(\Lambda, \theta)$ with respect to $\theta$ subject to the DE constraint~(\ref{constraint}). 
Introducing a Lagrange multiplier $\xi(t)$, we form the Lagrangian function
\[I(\Lambda, \theta, \xi)=G(\Lambda, \theta)-\int^{y_i}_0 \xi [\Lambda'(t)-h(\Lambda, t;x_i, \theta)]\dev t.\]
Because $\Lambda'(t)-h(\Lambda, t;x_i, \theta)=0$ for any $t$, the gradient of $G$ with respect to $\theta$ is equal to
\[\grad_{\theta}G=\frac{\partial I}{\partial \theta}=\int_{0}^{y_i}(1+\xi)(\frac{\partial h}{\partial \theta}+ \frac{\partial h}{\partial \Lambda}\frac{\partial \Lambda}{\partial \theta})\dev t -\int^{y_i}_0 \xi \frac{\partial \Lambda'}{\partial \theta}\dev t.\]
Using integration by parts, 
it follows that 
\begin{align*}
    \grad_{\theta}G & =\int_{0}^{y_i}(1+\xi)\frac{\partial h}{\partial \theta}\dev t \\
    & \ \ \ \ \ +\int_{0}^{y_i}\frac{\partial \Lambda}{\partial \theta}\left[\xi' +(1+\xi)\frac{\partial h}{\partial \Lambda}\right]\dev t - \left.\left(\xi \frac{\partial \Lambda}{\partial \theta}\right)\right|^{y_i}_0.
\end{align*}
Denote the adjoint $a(t)=\xi(t)+1$ and let $a(t)$ satisfy $a'(t)= - \frac{\partial h}{\partial \Lambda} a(t)$ and $a(y_i)=1$, then it follows that
$\grad_{\theta}G= \int_{0}^{y_i}a\frac{\partial h}{\partial \theta}\dev t.$
Calculation of the above integral requires the value of $\Lambda(t)$ and $a(t)$ along their entire trajectory from $0$ to $y_i$. Thus, we can compute the gradient $\grad_{\theta}G$ by solving the following augmented ODE which concatenates the dynamics and initial states of the three. Specifically, let $s(t) = [\Lambda(t), a(t), \grad_{\theta}G]$, then $s$ follows the ODE in (\ref{augode}).

\section{Partial Likelihood Based Methods}
\label{s: partial}

The Cox model~\citep{cox1972regression} and its extensions such as DeepSurv~\citep{katzman2018deepsurv} and Cox-Time~\citep{kvamme2019time} consider the hazard function in a semi-parametric way. Specifically, the conditional hazard function is factorized into two terms: a non-parametric baseline hazard function and a parametric relative risk function, that is \[\lambda_x(t) = h_0(t)\exp(g(t, x;\theta)).\]
The \textbf{Cox} model assumes a time-invariant linear relative risk where $g(t, x;\theta) = x^T\theta$. Subsequently, \textbf{DeepSurv} allows the relative risk to be a nonlinear function of feature $x$, i.e. $g(t, x;\theta) = g(x;\theta)$, but the proportional hazard assumption still holds; \textbf{Cox-Time} further allows the relative risk function to depend on time, which can handle the non-proportional hazard. In particular, DeepSurv and Cox-Time use neural networks to model $g(x;\theta)$ and $g(t, x;\theta)$. 

All the above models are fitted in two steps: the parameters in the relative risk function are learned through maximizing the partial likelihood function~\citep{cox1975partial}; the non-parametric cumulative baseline hazard function is obtained through the Breslow's estimator~\citep{breslow1972discussion} given the fitted relative risk in the first step.  

\paragraph{Partial likelihood.} The partial likelihood function is defined as
\[\mathcal{PL}(\theta;D) = \prod_{i:\Delta_i=1}\frac{\exp(g(y_i, x_i;\theta))}{\sum_{j\in R_i}\exp(g(y_i, x_j;\theta))},\]
where $R_i = \{j: y_j \geq y_i\}$ denotes the set of individuals who survived longer than the $i$-th individual, which is the so called \textit{at-risk} set. The estimator of $\theta$ is obtained by minimizing the negative log-partial likelihood function, that is 
\[\min_{\theta} \sum_{i:\Delta_i=1}[-g(y_i, x_i;\theta)+\log\sum_{j\in R_i}\exp(g(y_i, x_j;\theta))].\]

The partial likelihood function of each individual requires the access to the data of all individuals in the at-risk set. Hence, stochastic gradient decent (SGD) based algorithms cannot be directly applied. Although we can naively sample a mini-batch and restrict the at-risk set to individuals who are included in the current mini-batch in practice, there is a lack of theoretical justification. 

\paragraph{Breslow's estimator.} In order to obtain the predicted survival function, we need to estimate the cumulative hazard function. For models with the proportional hazard assumption, the estimated cumulative hazard function can be written as 
\[\hat{\Lambda}_x(t) = \int^{t}_0\hat{h}_0(s)\dev s \cdot \exp(g(x;\hat{\theta})) = \hat{H}_0(t)\exp(g(x;\hat{\theta})),\]
where $\hat{H}_0$ is the estimated cumulative baseline hazard function. The Breslow's estimator for $H_0$ is given by
\[\hat{H}_0(t) = \sum_{i:y_i\leq t}\frac{\Delta_i}{\sum_{j\in R_i}\exp(g(x_j;\hat{\theta}))}.\]
For Cox-Time with non-proportional hazard, the estimated cumulative hazard function is given by
\[\hat{\Lambda}_x(t) = \sum_{i:y_i\leq t}\frac{\Delta_i}{\sum_{j\in R_i}\exp(g(y_i, x_j;\hat{\theta}))}\exp(g(y_i,x;\hat{\theta})).\]
The survival function can then be estimated by $\hat{S}_x(t) = \exp(-\hat{\Lambda}_x(t))$.

\section{Discrete-Time Methods}
\label{s: discrete}
In the discrete-time setting, the range of possible values of the event time $T$ is divided into a set of disjoint intervals through pre-specified break points $\{t_0=0, t_1, \cdots, t_L\}$. Denote the intervals by $I_l = (t_{l-1}, t_l], l=1, \cdots, L$. Suppose the probability of occurrence of the event in time interval $I_l$ is $p_l (x) \geq 0$ with $\sum_{l=1}^Lp_l(x)=1$. The cumulative distribution $F_l$ and survival functions $S_l$ are, respectively
\[F_l(x) = \mathcal{P}\{T\leq t_l|X=x\} = \sum_{l=1}^l p_j(x),\ \  S_l(x)=\mathcal{P}\{T> t_l|X=x\}=1-F_l = 1-\sum_{j=1}^l p_j(x).\]
The conditional hazard probability $\lambda_l(x)$ is the probability that the event occurs in interval $I_l$ conditional on the survival up to the beginning of $I_l$, which could also determine the survival function through
\[\lambda_l(x) = \mathcal{P}\{T\in I_l|T\geq t_{l-1}, X=x\}=\frac{p_l(x)}{S_{l-1}(x)},\ \ S_l(x) = \prod_{j=1}^l (1-\lambda_j(x)).\]
Under the conditional independence assumption of the event time and the censoring time given features, the likelihood function is proportional to 
\[\prod_{i}p_{l_i}(x_i)^{\Delta_i}(1-\sum_{j=1}^{l_i-1}p_{j}(x_i))^{1-\Delta_i} = \prod_i [\lambda_{l_i}(x_i)\prod_{j=1}^{l_i-1}(1-\lambda_{j}(x_i))],\]
where $l_i$ is the index of time interval satisfying $t_{l_i-1}<y_i \leq t_{l_i}$.

\textbf{DeepHit}~\citep{lee2018deephit} models the probability mass function where the output of the neural network is a vector $[p_1(x), \cdots, p_L(x)]$. In addition to the negative log-likelihood (NLL) loss function, DeepHit considers another differentiable surrogate ranking loss tailored for time dependent concordance index, that is
\[\mathcal{L}_2=\sum_{i:\Delta_i = 1}\sum_{j:l_i<l_j}\eta(F_{l_i}(x_i), F_{l_i}(x_j)),\]
where $\eta(x,y)=\exp(\frac{-(x-y)}{\sigma})$ and $\sigma$ is a hyperparameter. They introduce another hyperparameter $\alpha$ to control the trade-off between the ranking loss and the NLL loss.  \textbf{Nnet-Survival}~\citep{gensheimer2018simple} models the conditional hazard probability where the output of the neural network is a vector $[\lambda_1(x), \cdots, \lambda_L(x)]$, and it is learned by maximizing the likelihood function.

\section{Hyperparameter Tuning}
\label{s: hyper}
We list the tuning ranges of hyperparameters for all neural network based models on three datasets in Table~\ref{table:hp}, where $\{\cdot\}$ represents the discrete search space and $[\cdot]$ represents the continuous search space\footnote{For the number of neurons, a real number is first sampled from the continuous space and then rounded to the closest integer.}. Specifically, we tune the rate of dropout and batch normalization for DeepSurv, DeepHit, Nnet-Survival, and Cox-Time. We treat the number of time intervals as a hyperparameter for DeepHit and Nnet-Survival. We also tune two hyperparameters, $\alpha$ and $\sigma$, associated with the surrogate ranking loss in DeepHit. Since the three datasets are of different sizes, we use different search ranges for the batch size: $\{32, 64, 128, 256\}$ for METABRIC, $\{128, 256, 512\}$ for SUPPORT, and $\{512, 1024\}$ for MIMIC and MIMIC-SEQ. The discrete models (DeepHit and Nnet-Survival) appear to be sensitive to the number of time intervals on different datasets. Therefore we search the number of time intervals for these two discrete models from $\{10, 50, 100, 200, 400\}$ for the smaller datasets, METABRIC and SUPPORT, and from $\{50, 100, 200, 400, 800\}$ for the larger datasets, MIMIC and MIMIC-SEQ.

\begin{table}[!h]
	\centering
	\begin{tabular}{l|c}
		\toprule
		Number of dense hidden layers & $\{1, 2, 4\}$\\
		Number of neurons in each dense hidden layer & $[2^2, 2^7]$\\
		Number of neurons in each GRU hidden layer & $[2^3, 2^8]$\\
		Learning rate & $[10^{-4.5}, 10^{-1.5}]$\\
		Weight decay & $[10^{-9}, 10^{-4}]$\\
		Momentum & $[0.85, 0.99]$\\
		Dropout (DeepHit, DeepSurv, Nnet-Survival, Cox-Time) & $\{0, 0.1, 0.5\}$\\
		Batch normalization (DeepHit, DeepSurv, Nnet-Survival, Cox-Time) & $\{\text{True}, \text{False}\}$\\
		$\alpha$ (Surrogate ranking loss in DeepHit) & $[0, 1]$\\
		$\sigma$ (Surrogate ranking loss in DeepHit) & $\{0.25, 1, 5\}$\\
		\bottomrule
	\end{tabular}
	\caption{Tuning ranges of hyperparameters for experiments on the real-world datasets.}
\label{table:hp}
\end{table}

\end{document}